\pdfoutput=1

\documentclass[11pt]{article}

\usepackage[final]{acl}

\usepackage{times}
\usepackage{latexsym}
\usepackage{amsmath, fge}
\usepackage{comment}

\usepackage[T1]{fontenc}

\usepackage[utf8]{inputenc}

\usepackage{microtype}

\usepackage{inconsolata}
\usepackage{tabularx}

\usepackage{graphicx}

\usepackage{todonotes}
\newcounter{todocounter}

\usepackage{makecell, cellspace, caption}
\usepackage[table, svgnames, dvipsnames]{xcolor}
\usepackage{amsmath}
\usepackage{amssymb}
\usepackage{multirow}
\usepackage{caption}
\usepackage{subcaption}
\usepackage{graphicx}
\usepackage{float}
\usepackage{adjustbox}
\usepackage{graphicx}
\usepackage{svg}
\usepackage{booktabs}
\usepackage{xcolor,colortbl}
\usepackage{enumitem}

\title{Confidence-Based Response Abstinence: Improving LLM Trustworthiness via Activation-Based Uncertainty Estimation}

\author{
 \textbf{Zhiqi Huang\textsuperscript{\( \dag  \)}},
 \textbf{Vivek Datla}\textsuperscript{\( \dag\ddag \)}, 
 \textbf{Chenyang Zhu},
 \textbf{Alfy Samuel},\\
 \textbf{Daben Liu},
 \textbf{Anoop Kumar},
 \textbf{Ritesh Soni}
 \\
 Capital One \\
\texttt{\{zhiqi.huang, vivek.datla, chenyang.zhu, alfy.samuel,} \\
\texttt{daben.liu, anoop.kumar, ritesh.soni\}@capitalone.com}\\
\small{
 \textsuperscript{\( \dag  \)}Equal Contribution,  \textsuperscript{\( \ddag  \)}Corresponding Author
}
}

\begin{document}

\maketitle

\begin{abstract}
We propose a method for confidence estimation in retrieval-augmented generation (RAG) systems that aligns closely with the correctness of large language model (LLM) outputs. Confidence estimation is especially critical in high-stakes domains such as finance and healthcare, where the cost of an incorrect answer outweighs that of not answering the question. Our approach extends prior uncertainty quantification methods by leveraging raw feed-forward network (FFN) activations as auto-regressive signals, avoiding the information loss inherent in token logits and probabilities after projection and softmax normalization. We model confidence prediction as a sequence classification task, and regularize training with a Huber loss term to improve robustness against noisy supervision. Applied in a real-world financial industry customer-support setting with complex knowledge bases, our method outperforms strong baselines and maintains high accuracy under strict latency constraints. Experiments on Llama 3.1 8B  model show that using activations from only the 16th layer preserves accuracy while reducing response latency. Our results demonstrate that activation-based confidence modeling offers a scalable, architecture-aware path toward trustworthy RAG deployment.
\end{abstract}

\section{Introduction}

In high-stakes applications like financial customer support, it is often more desirable and trustworthy for a Retrieval Augmented Generation (RAG) system to abstain from answering than to risk providing an incorrect response. Although not responding to a query reduces the system's immediate utility, it is a necessary trade-off to ensure accuracy and preserve user trust. The guiding principle is that the reputational and financial cost of providing a wrong answer is significantly higher than the cost of not providing one. This challenge requires a principle of abstention.

\begin{figure*}[!htbp]
\centering
\includegraphics[width=0.95\textwidth]{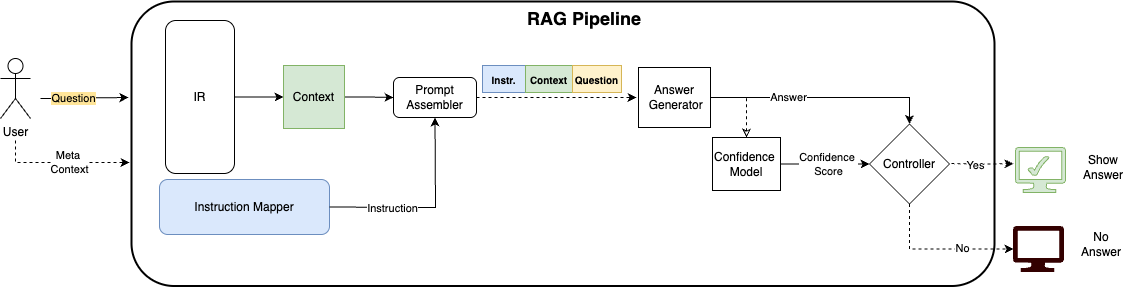}
\caption{Diagram of the proposed Retrieval Augmented Generation (RAG) with the confidence model. When a user asked a question, the IR component retrieves related context from a database. The prompt is then constructed and sent into a question and answering LLM. A confidence score would be generated by the confidence model and being used to control whether or not to show the result to the user. }
\label{fig:utility}
\end{figure*}
One way to achieve the abstention is to have a confidence measure that correlates with correctness of the response, and mask the response when the confidence score is below a threshold. Uncertainty of the model while generating the response is a viable source of signal for building a confidence measure. 

To develop a practical solution, it is crucial to identify the primary source of this uncertainty. In highly regulated fields, the error is rarely due to aleatoric uncertainty (randomness inherent in the data), as knowledge bases are typically vetted by legal and subject-matter experts. The more probable source is epistemic uncertainty (the model's own lack of knowledge), which arises when the model's parametric knowledge, acquired during pre-training or fine-tuning, conflicts with or misinterprets the provided context.

While existing approaches ~\cite{bakman2024mars,liu2024uncertainty, malinin2020uncertainty, kuhn2023semantic} to uncertainty estimation in retrieval-augmented generation (RAG) have shown promise, they often fall short when the target response is long and narrative in nature. This challenge becomes especially pronounced in sensitive domains such as finance, where queries can be ambiguous or underspecified. For instance, a question like "What is the deadline to make a payment on Card Type A?" may retrieve multiple similar documents, each corresponding to different subcategories of the card type. In such cases, both the query and the retrieved context exhibit ambiguity, which can propagate through the RAG pipeline. Simply measuring uncertainty based on generated response is insufficient to ensure correctness.

Also, methods relying on sampling \cite{bakman2024mars}, are less practical at scale. These techniques rely on generating a response multiple times with slight variations to measure the model's consistency, a process that introduces prohibitive computational costs and latency in a production environment. For RAG systems that must serve users in real-time, such multi-generational approaches are not a viable solution.

Uncertainty and correctness, while related, are fundamentally distinct concepts \cite{liu2025uncertainty}. A model's low uncertainty in its output does not necessarily imply correctness, just as a model may generate a correct response with a high uncertainty. This distinction becomes particularly salient in retrieval-augmented generation (RAG) applications, where correctness often hinges on factual grounding rather than surface-level fluency. Our goal is to utilize the model's internal uncertainty signals to generate a confidence score that correlates strongly with the correctness of the response generated by an LLM. 

We build our confidence model using the raw activation signals inside the feedforward layers of LLM which include the activations of knowledge neurons~\cite{azaria-mitchell-2023-internal}. Thus, our model captures the relationship between the auto-regressive properties of activations and inherent uncertainty of the model in generating a response. 
We propose a supervised framework to train a sequence classifier model and  generate a confidence score that correlates with response correctness.


Figure~\ref{fig:utility} illustrates the practical utility of integrating a confidence model into our RAG pipeline. The primary goal of the system is to provide users with accurate answers. However, in cases where there is insufficient epistemic or aleatoric knowledge to reliably answer a question, the system's next best action is to abstain from answering. This behavior is enabled by a controller that filters responses based on their confidence scores, allowing the system to avoid potentially incorrect or misleading outputs. This system is deployed in production for large-scale use that achieves high precision  while maintaining an acceptable display rate (defined as the ratio of response pass the confidence filter to total responses generated by the system). Experimental results show that our confidence model outperforms multiple baselines, reaching a precision of 0.95 with 70.1\% display rate (masking 29.9\% of the total responses). Furthermore, when compared to ground truth, displayed responses exhibit a significantly higher ROUGE score than masked responses.


\section{Related Work}

\begin{figure}[h]
\centering
\includegraphics[width=0.45\textwidth]{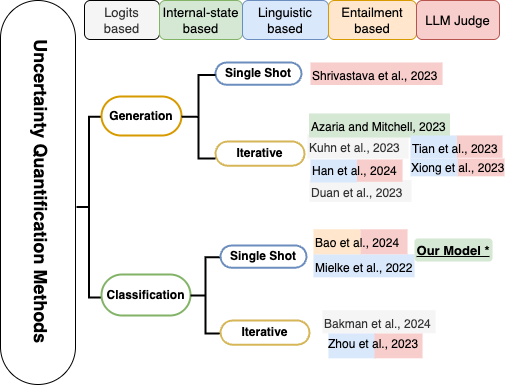}
\caption{Landscape of Uncertainty Quantification Methods}
\label{fig:realted_work}
\end{figure}

Figure~\ref{fig:realted_work} shows the landscape of various uncertainty quantification methods in LLMs. When mapping the landscape, they can be broadly grouped by the strategies used to quantify the uncertainty. 

\citet{shrivastava2023llamas} demonstrate that the generation probabilities of weaker white-box models (that is, smaller models) can be used to estimate the internal confidence levels of larger black-box models. The approach involves zero-shot generation using prompt variations based on different instructions to infer the confidence of responses produced by the larger model. \citet{duan2023shifting} and \citet{kuhn2023semantic} use semantic entropy to reweight token-level importance, prioritizing content-bearing tokens while discounting filler words. Their core intuition is that if semantically important tokens are generated with high confidence, the overall response is more likely to be correct even if less important tokens exhibit lower confidence. 

\citet{azaria-mitchell-2023-internal} show that the LLMs internal parameters show tell-tale signs when generating text with uncertainty. When a model's generation path falls into a speculative region, evidenced by competition between two or more plausible next tokens, its confidence is adversely affected. They introduce small input perturbations to induce trajectory shifts, and monitor corresponding changes in token-generation activations and outputs. They label the speculative generation as lying, and propose that activation patterns can shed light on this speculative generation. This method requires white-box access to the model to obtain token-level probability traces.

\citet{tian2023just} have empirically shown that LLM's self generated confidence score while giving a response could be calibrated by sampling over perturbed questions. Specifically, they show that prompting a model to produce several answer choices before giving its confidence scores  helps in calibration of verbalized probabilities.

In a related direction, \citet{xiong2023can} generate multiple variants of a prompt using diverse prompting strategies such as Chain-of-Thought (CoT), self-probing, and top-k sampling. They utilize a separate LLM as a "judge model" to evaluate each variant and assign a confidence score. Variations in these scores are then used to predict the confidence of the target model's original response. Similarly, \citet{han2024enhancing} proposed a confidence measurement based on the perturbation of the question. The variation in model's answer generation probabilities for various perturbations of the question for the same context is used as a measure to generate a verbalized confidence score.

Several recent studies adopt a classification-based approach to estimate response plausibility, offering a more computationally efficient alternative by avoiding multiple generations. For example, HHEM~\cite{hhem-2.1-open} uses an entailment-based model to measure the semantic coherence between the input and the generated output. This approach operates under black-box constraints, requiring only the input-output pair from the target LLM to assess the correctness of the response.

Other methods focus on linguistic cues as indicators of ambiguity in LLM outputs. \citet{mielke2022reducing} argue that model confidence does not always correlate with correctness and show that linguistic calibration of input prompts can significantly influence a model's confidence. They introduce a calibration score that helps generate more accurate responses by aligning linguistic features with expected confidence levels. Their evaluations were performed on factoid QA datasets, where there is a zero-sum approach towards correctness. We argue that when the parametric knowledge of the LLM is mainly contributing to the style of the response, and the key facts come from the input, confidence can serve as an effective signal for correctness.

Our method draws inspiration from prior work on activation-based knowledge tracing~\cite{dai-etal-2022-knowledge}, generation trajectory modeling~\cite{azaria-mitchell-2023-internal}, and importance-weighted token probabilities~\cite{bakman2024mars}. \citet{dai-etal-2022-knowledge} highlight how feedforward network (FFN) activations encode key factual information, showing that the activation of certain neurons is positively correlated with knowledge expression. Building on this insight, we treat FFN activations as autoregressive signals and train a recurrent neural network (RNN) to predict the probability that a model-generated response is correct. A score closer to 1 indicates greater model confidence in the response's correctness.

\section{Method}


For a generated response sequence  $s$  of length $L$ for the given input $x$ to a model $M$ with parameters $\theta$, the probability of generating the sequence is given as follows:
\begin{equation}
P(s \mid x; \theta) = \prod_{l=1}^{L} P(s_l \mid s_{<l}, x; \theta)
\end{equation}

To compare sequence probability across different lengths of generated output, previous approaches have normalized the score based on the length of the response. The length‑normalized score, used in prior uncertainty estimation (UE) methods ~\cite{malinin2020uncertainty}: 
\begin{equation}
\tilde P(s \mid x; \theta) \;=\; \left( \prod_{l=1}^L P(s_l \mid s_{<l}, x; \theta)\right)^{1/L}
\end{equation}

Here all the tokens contribute are given equal importance irrespective of the length of sequence. The risk with this approach is that a single low-probability unusual word can disproportionately lower the overall sequence score, even if subsequent tokens have high probabilities.

Several of the methods that perform uncertainty estimation taking token-logits perform similar weighing and they have shown great results in factoid question answering. These methods do not scale for longer answers, where there are multiple sentences and few tokens don't hold the key to correctness. Also, multiple generations needed to quantify the confidence score make them prohibitively expensive in a large scale settings.


\begin{figure}[!htbp]
\centering
\includegraphics[width=0.5\textwidth]{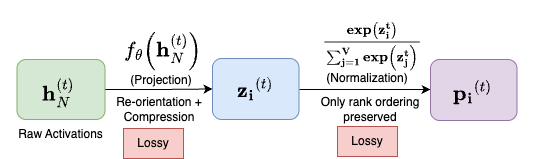}

\caption{Motivation to use activations instead of token probabilities.}
\label{fig:lossy_signal}
\end{figure}

Our goal is to estimate the correctness of the generated response in a single shot using uncertainty estimation. We prefer using FF-layer activations rather than token probabilities because token probabilities are computed by applying the decoder head (a linear projection) followed by a softmax transformation. This projection compresses the rich internal representation into a vocabulary space and the softmax operation further distorts the signal by normalizing it into a probability distribution (see Figure~\ref{fig:lossy_signal}), potentially obscuring fine-grained differences in the model's internal state. In contrast, raw activations preserve the high-dimensional representation prior to this compression, providing a more direct view of the model's internal dynamics during response generation.

\subsection{Our Confidence Model}

\begin{figure*}[htbp]
\centering
\includegraphics[width=1.0\textwidth]{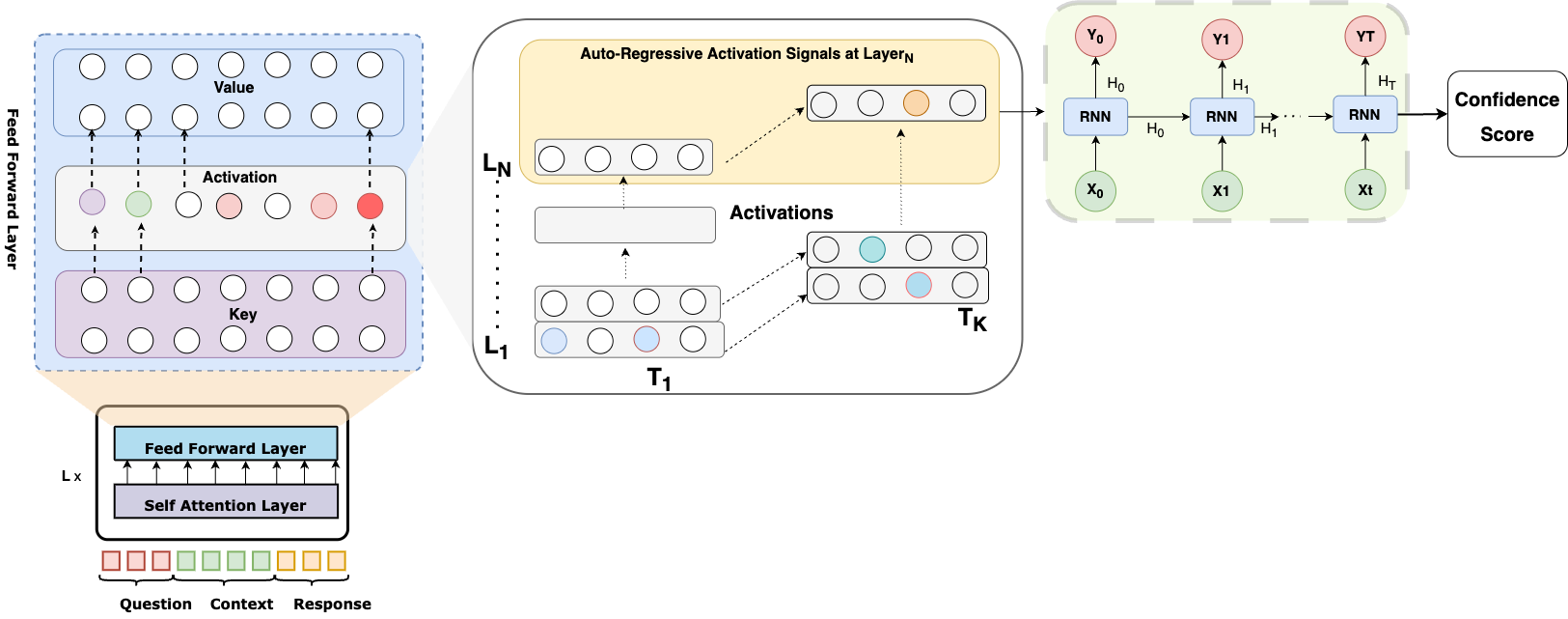}

\caption{Confidence model based on the activations of large language models. Our method first feed the <Question, Context, Response> pair in an LLM. We then extract the activations from the 32-th or 16-th layer, and feed the activations into an LSTM and a classification head. The classification logit serves as the confidence score. }
\label{fig:arp_model}
\end{figure*}


Figure~\ref{fig:arp_model} shows a graphical representation of our confidence model. To estimate the confidence of a generated answer $s$ of size $L$, we introduce a lightweight, trainable probe that operates on the internal representations of the Llama 3.1 8B model. The process begins by providing a structured prompt to the LLM, which is formulated as a sequence of tokens, $x$ of size $T+L+1$, which is a concatenation of the following: Instruction($x_I$), Question($x_Q$) and Context($x_C$) of size $T$ tokens; Answer($s$) of size $L$ tokens; and EOS token($x_{EOS}$) of size $1$.
The complete input sequence is formally represented as:
\begin{equation}
x = x_I \oplus x_Q \oplus x_C \oplus s \oplus x_{EOS}    
\end{equation}
 where $\oplus$ denotes the concatenation operation.

During a single forward pass through the LLM, we extract the hidden state activations from a specific transformer layer, $\ell$. We investigate representations from two distinct depths within the network: the final layer ($\ell=32$) and a middle layer ($\ell=16$). This yields a full sequence of hidden state vectors

\vspace{-10pt}
\begin{equation}
    \mathbf{H}_\ell = (\mathbf{h}_{\ell}^{1}, \dots, \mathbf{h}_{\ell}^{T+L+1})
\end{equation}

Each vector $\mathbf{h}_{\ell,k} \in \mathbb{R}^{d_{\text{LLM}}}$ corresponds to the $k$-th input token, with size of LLM's activation dimension. From this complete set of activations, we isolate only those corresponding to the tokens of the candidate answer, which span from index $T+1$ towards the final $x_{EOS}$ token. This forms the input sequence, $S_{\text{in}}$, for our confidence estimation module:

\vspace{-10pt}
\begin{equation}
S_{\text{in}} = (\mathbf{h}_{\ell}^{T+1}, \mathbf{h}_{\ell}^{T+2}, \dots, \mathbf{h}_{\ell}^{T+L+1})
\end{equation}

The extracted sequence $S_{\text{in}}$ is then fed into a sequence classifier $g(S_{in})$, which is trained to model the sequence of activations. The sequence classifier with a classification head outputs a 2-dimensional logit vector, $\mathbf{z}$, such that the confidence score can be computed as, 
\begin{equation}
    c = \text{softmax}(\mathbf{z})_1 = \frac{e^{z_1}}{e^{z_0} + e^{z_1}}
\end{equation}

Our goal is to estimate the confidence of the model when generating an answer, with ulterior goal of rejecting the generated answer if $c$ falls below a threshold of confidence. In this framework, only the parameters of the sequence classifier $g(S_{in})$ are trainable. We use a Long short-term memory (LSTM)~\cite{sutskever2014sequence} as the sequence classifier for the following experiments.

\subsection{Model Training}
Given that the retrieval stage of the pipeline may introduce alethic knowledge gaps, the input context provided to the LLM can be incomplete, or contain contradictory information across the document retrieved. To address this, we introduce an explicit regularizer based on Huber loss $L_{\text{Huber}}$, which is more robust to such noise~\cite{patra2023calibrating}. Unlike just using only the Cross-Entropy loss $L_{\text{CE}}$, which can be highly sensitive to large deviations when predictions are far from the target, the Huber loss based regularizer helps smoothen with a linear penalty for large errors. This property reduces the influence by outliers arising from imperfect retrieval.

\vspace{-10pt}
\begin{equation}
H_\delta(x) =
\begin{cases} 
\frac{1}{2} x^2 & \text{for } |x| \leq \delta \\
\delta \left( |x| - \frac{1}{2} \delta \right) & \text{otherwise}
\end{cases}
\end{equation}

where $\delta > 0$ is a hyperparameter that controls the transition point between the quadratic and linear loss. 


Using $L_{CE}$ loss with $L_{Huber}$ regularizer, we learn to predict confidence score, which correlates with the correctness. The higher the confidence, the higher are the changes for the generated output to be correct. For a sampled minibatch $ B = \{ (x_j, y_j) \}_{j=1}^{|B|} $, the Huber loss term is calculated as:
\vspace{-10pt}
\begin{equation}
L_{\text{Huber}} = H_\delta \left( \frac{1}{|B|} \sum_{i=1}^{|B|} c_i - \frac{1}{|B|} \sum_{i=1}^{|B|} I(\hat{y}_i = y_i) \right)
\end{equation}
where \( c_i = \max(\hat{y}_i) \) is the confidence of the prediction for instance \( x_i \), and \( I(\hat{y}_i = y_i) \) is the indicator function for correct predictions.

The total loss function 
\begin{equation}
L_{Total} = L_{CE} + \lambda L_{Huber}
\end{equation} where $\lambda$ controls the strength of regularization.

In our modeling, several constraints arise naturally from the real-time conditions under which the system operates. The generated output must remain grounded in the input context provided within the prompt. The output must adhere to predefined stylistic or structural patterns required to present certain types of information. At the end of generation, an explicit decision signal determines whether the answer is shown to the user. This signal is conditioned on multiple factors, including:
\begin{itemize} 
\item Subject-matter-expert (SME) defined standards of correctness for the class of questions.
\item The requirement that factual content be derived from the input context, while stylistic elements may rely on the model’s parametric knowledge.
\end{itemize}
We conducted experiments on our proprietary knowledge corpus consisting of procedures, rules, and complex instructions to be followed to address the various needs of support agents handling a large volume of customer base. Our results indicate a robust performance using our method compared to the several SOTA UQ and hallucination identification methods.

\section{Experimentation}

We have conducted experiments to identify the optimal masking ratio in order to maintain utility and precision of the system.

\subsection{Data}
\subsubsection{Disclosure on data}
Due to the sensitive nature of the data, which pertains to proprietary financial tools and internal knowledge resources used by service agents within a financial institution, we are unable to share dataset details. This restriction is in place to ensure compliance with internal data governance policies and to protect confidential and regulated financial information. We hope that the community understands the importance of maintaining the integrity and privacy of such sensitive operational data.

\begin{figure*}[htbp]
\centering
\includegraphics[width=0.95\textwidth]{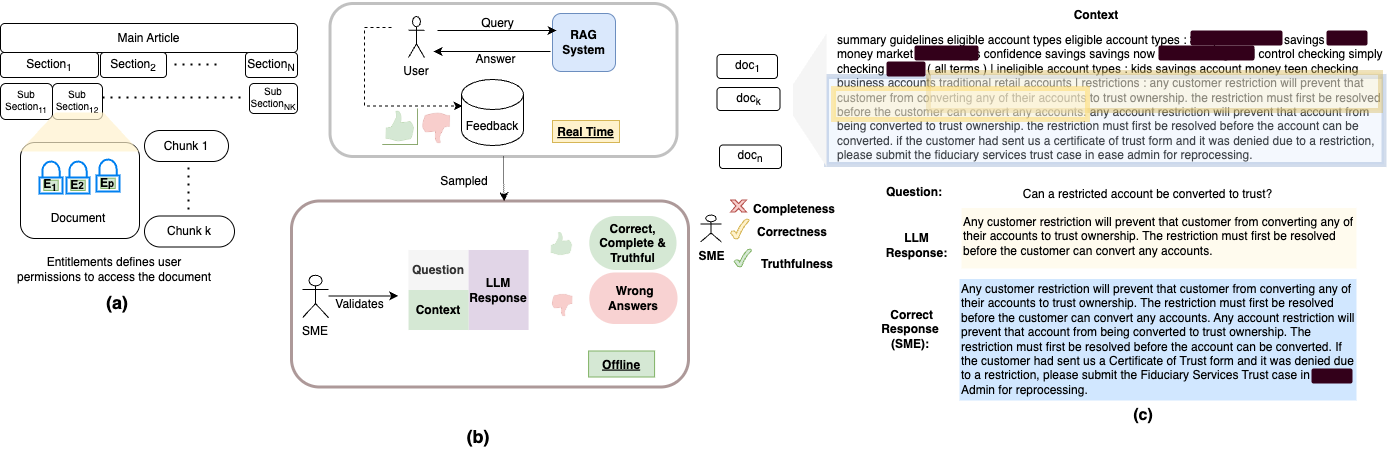}

\caption{Features of our Knowledge base.  (a) Complex structure of our knowledge articles; (b) Process of SME validated training data creation; (c) Example SME validated LLM Response}
\label{fig:combined_data_features}
\end{figure*}

\subsubsection{Features of our knowledge articles}
We provide an overview of the population-level characteristics of our dataset, which is derived from a knowledge base composed of instructional articles designed to guide customer support agents in using proprietary internal tools. These tools are governed by strict procedural guidelines essential for resolving customer issues. For instance, when handling a customer inquiry about a specific transaction, agents must follow a prescribed sequence: verifying the customer's identity, obtaining consent to access the account, identifying the relevant transaction, and initiating additional processes such as flagging the transaction in cases of suspected fraud. 

Figure~\ref{fig:combined_data_features}(a) shows the hierarchical nature of the documents. Our knowledge base is semi-structured comprising heterogeneous documents with rich hierarchical and content structures. These documents may include deeply nested sections (e.g., sections, subsections, sub-subsections), as well as complex content types such as tables, bullet and numbered lists, and embedded entities. 

Each subsection article is treated as a separate document. Each document is further chunked to be efficiently indexed in a low latency store. Overall, there are $8.5k$ documents and approximately $45k$ chunks in the knowledge-base. 

\subsubsection{Features of the training data}

Our system design incorporates a real-time feedback loop, as illustrated in Figure~\ref{fig:combined_data_features}(b), where support agents interact with the RAG system and provide immediate feedback (e.g., thumbs-up/down) on the usefulness of generated responses. Processing thousands of these interactions daily, we draw a stratified sample of both positive and negative feedback instances, accounting for dimensions like product type and line of business. For each sampled case, we collect the query, generated answer, retrieved context, and associated metadata for a more rigorous offline evaluation.

This offline review is conducted by subject matter experts (SMEs) who assess each answer for completeness, correctness, and truthfulness, ensuring it is grounded in the provided context rather than inferred from the model's parametric knowledge. SMEs may also refine responses to create ideal, complete answers, as shown in the example in Figure~\ref{fig:combined_data_features}(c). This two-tiered approach of combining real-time user signals with deep SME validation allows us to build a high-quality labeled dataset for training and evaluation, ensuring the model aligns with domain-specific requirements for accuracy and trustworthiness.

\subsection{Information Retrieval}

We perform retrieval using an open-search index configured for K-nearest neighbor (KNN) retrieval based on semantic similarity to the input query. In addition to the query itself, we incorporate associated metadata such as entitlements and access-control filters specific to the agent submitting the question, to ensure that the retrieved documents adhere to the agent's permissions.

In the context of this work, we do not explicitly quantify retrieval errors. Instead, our focus lies in modeling the generation process of the response. We assume the retrieval step to be correct and treat errors introduced during retrieval as alethic uncertainty, while the knowledge encoded within the model through pretraining and fine-tuning is considered epistemic. Our confidence model is designed to map the relationship between the question, the retrieved (alethic) knowledge, the model’s internal (epistemic) knowledge, and the generated response. This relationship is captured through patterns in the model’s internal activations, treated as auto-regressive signals.

We observe that this mapping cannot be adequately modeled using a simple feedforward (MLP) architecture, as it fails to capture the temporal dependencies inherent in the generation process. Therefore, we adopt a recurrent architecture specifically, a lightweight Long Short-Term Memory~\cite{hochreiter1997long} (LSTM), trained using $L_{CE}$ loss and $L_{Huber}$ regularizer loss. The LSTM is trained on input sequences derived from the activations of a selected layer, along with carefully curated training data that aligns the activation patterns with response-level confidence.

\begin{table}[!htbp]
\centering
\begin{adjustbox}{width=0.3\textwidth}
    \begin{tabular}{l|c}
    \hline
    Method & AUROC \\
    \hline
    Vectara & 0.590 \\
    $\text{Vectara}_{\text{FT}}$ & 0.634 \\
    $\text{Logits}_{\text{based}}$ & 0.663 \\
    $\text{Our Model}_{\text{no calib.}}$ & 0.741\\
    \rowcolor{LightBlue}$\textbf{Our Model}_{\textbf{with calib.}}$ & \textbf{0.772} \\
    \hline
    \end{tabular}
\end{adjustbox}
\caption{Comparing our approach to other baselines}
    \label{res:baseline_compare}
\end{table}

\begin{table}[!htbp]
\centering
\begin{adjustbox}{width=0.48\textwidth}
    \begin{tabular}{c|cc|cc|c}
    \hline
    \multirow{2}{*}{Threshold} & \multirow{2}{*}{P} & \multirow{2}{*}{R} & \multicolumn{2}{c|}{ROUGE-L} & \multirow{2}{*}{\%Mask} \\
    & & & Display & Mask &  \\
    \midrule
    Baseline (0.0) & 0.90 & 1.00 & 0.62 & N/A & 0.0 \\
    \midrule
    0.1	& 0.94	& 0.89  & 0.64  & 0.54  & 14.4 \\
    0.2	& 0.94	& 0.83	& 0.64	& 0.56	& 20.4 \\
    0.3	& 0.94	& 0.80	& 0.64	& 0.57	& 22.9 \\
    \rowcolor{LightBlue} 0.4	& 0.94	& 0.76	& 0.64	& 0.57	& 26.4 \\
    \rowcolor{LightBlue} 0.5	& 0.95	& 0.73	& 0.65	& 0.57	& 29.9 \\
    \rowcolor{LightBlue} 0.6	& 0.95	& 0.69	& 0.66	& 0.56	& 34.8 \\
    0.7	& 0.96	& 0.65	& 0.66	& 0.57	& 38.6 \\
    0.8	& 0.96	& 0.60	& 0.66	& 0.58	& 44.0 \\
    0.9	& 0.97	& 0.52	& 0.67	& 0.58	& 52.0 \\
    \bottomrule
    \end{tabular}
    \end{adjustbox}
    \caption{Our Confidence score model with calibration helps achieve 0.95 precision while masking 29.9\% of the responses}
    \label{res:best_scores}
\end{table}

\begin{table}
\centering
\begin{adjustbox}{width=0.45\textwidth}
    \begin{tabular}{c c c c c c}
    \hline
     \multirow{2}{*}{IR Model}& R@1 & R@3 & R@5 & R@10 & R@25 \\
    \cmidrule{2-6}
     & 0.54 & 0.75 & 0.80 & 0.84 & 0.88 \\
    \hline
    \end{tabular}
\end{adjustbox}
\caption{Current recall(r) of the IR system, that helps in creating the context for the RAG pipeline}
    \label{res:ir_scores}
\end{table}

\subsection{Results}

\begin{table}[t]
\centering
\begin{adjustbox}{width=0.48\textwidth}
    \begin{tabular}{cc|cc|cc|c}
    \hline
    \multirow{2}{*}{Layer} & \multirow{2}{*}{Context} & \multirow{2}{*}{P} & \multirow{2}{*}{R} & \multicolumn{2}{c|}{ROUGE-L} & \multirow{2}{*}{\%Mask} \\
    & & & & Display & Mask & \\
    \hline
    32 & Full & 0.95 & 0.73 & 0.65 & 0.57 & 29.9\\
    32 & Top 5 & 0.95 & 0.69 & 0.66 & 0.56 & 34.3\\
    32 & Top 3 & 0.96 & 0.63 & 0.66 & 0.57 & 40.5\\
    32 & Top 1 & 0.97 & 0.56 & 0.67 & 0.57 & 47.5\\
    \midrule
    16 & Full & 0.97 & 0.73 & 0.64 & 0.58 & 31.3 \\
    16 & Top 5 & 0.98 & 0.65 & 0.65 & 0.59 & 39.3 \\
    16 & Top 3 & 0.98 & 0.60 & 0.66 & 0.58 & 44.8 \\
    16 & Top 1 & 0.99 & 0.48 & 0.66 & 0.59 & 56.2 \\
    \bottomrule
    \end{tabular}
    \end{adjustbox}
    \caption{Identifying the optimal  setting to run confidence model}
    \label{res:optimizing_context_size_layers}
\end{table}

Our method achieves superior calibration of LLM responses, maintaining high precision with minimal utility loss. As shown in Table~\ref{res:baseline_compare}, it outperforms industry SOTA methods, Vectara (HHEM2.1)~\cite{hhem-2.1-open} and a logits-based uncertainty model~\cite{malinin2020uncertainty}. We obtain further performance gains  by caliberating with $L_{\text{Huber}}$ as a regularizer.

Table~\ref{res:best_scores} reports confidence thresholds that optimize precision while keeping the masking rate low. Although an ideal mask rate is $0\%$, realistic applications must tolerate some masking due to noise in LLM inputs. In our setup, the retrieval stage achieves a strong $recall@10 > 0.8$ (Table~\ref{res:ir_scores}), yet residual alethic knowledge gaps in retrieval can affect downstream generation.

We experimented with varying input context sizes, selecting the top $k$ documents ($k \in \{1, 3, 5, 7 \ \text{(full)}\}$), and with partial-layer activation extraction from Llama 3.1 8B (layer 16 or layer 32)~\cite{llama3modelcard}. As shown in Table~\ref{res:optimizing_context_size_layers}, using activations from only the 16th layer yields performance on par with the full-layer setup while maintaining a reasonable mask rate.

\begin{table}[t]
\centering
\begin{adjustbox}{width=0.38\textwidth}
\renewcommand*{\arraystretch}{1.1}
    \begin{tabular}{ccc|c|c}
    \hline
    Framework & Layer & Context & Avg. ms & P99 \\
    \hline
    \multirow{5}{*}{Hugging Face} & \multirow{4}{*}{32} & Full & 221 & 387 \\
    & & Top 5 & 179 & 329 \\
    & & Top 3 & 137 & 286 \\
    & & Top 1 & 100 & 252 \\
    \cmidrule{2-5}
    & 16 & Full & 139 & 278 \\
    \hline
    \multirow{5}{*}{vLLM} & \multirow{4}{*}{32} & Full & 206 & 354 \\
    & & Top 5 & 161 & 304 \\
    & & Top 3 & 125 & 269 \\
    & & Top 1 & 88 & 241 \\
    \cmidrule{2-5}
    & 16 & Full & 127 & 267 \\
    \bottomrule
    \end{tabular}
\end{adjustbox}
\caption{Latency of the confidence model using various context sizes, Avg. time is calculated across 3 runs of the same input.}
    \label{res:latency}
\end{table}

Latency analysis (Table~\ref{res:latency}) confirms that input context size is a dominant factor; larger contexts increase response time, highlighting a trade-off between context size and system responsiveness. In the production system, the confidence model is deployed with vLLM \cite{kwon2023efficient}, and overall the same trend appears there as well.

\section{Discussion}

In this work, we present an approach for constructing a confidence score that aligns with the correctness of responses generated by large language models (LLMs). Such a measure is particularly critical in high-stakes domains such as finance and healthcare, where the cost of an incorrect response far exceeds that of withholding a response. Our method extends prior works in uncertainty quantification (UQ) \cite{malinin2020uncertainty, hhem-2.1-open} by leveraging model activation patterns to predict correctness more robustly.

Figure~\ref{fig:lossy_signal} illustrates our motivation for using raw activation signals from the feed-forward network (FFN) layers as auto-regressive features, rather than token logits or probabilities. Token probabilities are obtained after a linear projection and softmax transformation. The projection step reduces dimensionality, discarding non-vocabulary-aligned features, while the softmax normalization saturates probability values, erasing scale information and compressing relative differences. Using activations directly, we retain the full representational capacity of the internal state of the model.

Our application setting involves customer support agents consulting a proprietary knowledge base to resolve customer queries using specialized internal tools. The knowledge base contains documents vetted across multiple dimensions, including risk and legal compliance, making factual errors in the content highly unlikely. However, strict permissions govern which documents an agent can access. Figure~\ref{fig:combined_data_features}(a) shows the complexity of document formats and fine-grained entitlements that impact retrieval and downstream generation.

We model confidence estimation as a classification problem over sequences of activations. Specifically, we employ a lightweight recurrent neural network (LSTM) that consumes FFN activations as auto-regressive signals. The classification logit from the LSTM head serves as the confidence score (see Figure~\ref{fig:arp_model}). To enhance robustness against noisy supervision, we introduce a Huber loss regularizer $L_{\text{Huber}}$ alongside the cross-entropy loss $L_{\text{CE}}$. The Huber loss's ability to behave quadratically for small errors and linearly for large errors makes it well-suited for smoothing gradients and mitigating the influence of outliers~\cite{patra2023calibrating}. Results in Table~\ref{res:baseline_compare} demonstrate that our approach outperforms strong baselines, and the inclusion of $L_{\text{Huber}}$ further improves accuracy over using $L_{\text{CE}}$ alone.

In real-world deployment, retrieval-augmented generation (RAG) pipelines must meet strict latency requirements, as the LLM prompt length is constrained by model context limits and thousands of queries are processed daily. Tables~\ref{res:optimizing_context_size_layers} and \ref{res:latency} summarize our performance–latency trade-offs. Reducing the number of Llama 3.1 8B layers from 32 to 16 while keeping context size fixed preserves accuracy while reducing latency by approximately 42.5\%. When the context size is reduced, alethic errors increase due to incomplete retrieval, raising the model’s masking rate (i.e., instances where no answer is returned due to low confidence). Nevertheless, the 16-layer configuration achieves comparable performance to the 32-layer setup at lower computational cost. We observe a slight improvement in response latency when hosting the model using vLLM inference compared to Hugging Face's inference API, likely due to vLLM's optimized memory management and continuous batching capabilities.

Overall, our approach leveraging FFN activations as auto-regressive signals, modeling them with an LSTM, and regularizing with $L_{\text{Huber}}$ proves effective in long-form RAG settings. This method improves the trustworthiness of LLM-generated responses and holds strong potential for safe deployment in sensitive, domain-specific applications.

\section{Limitations}

Our work pushes the boundary of confidence estimation in retrieval-augmented generation (RAG) for sensitive domains, but several practical considerations remain. Ideally, a RAG system should generate both the response and its confidence score in a single pass. In our current implementation, the confidence score requires a second run of the system, which introduces additional computational and latency overhead. 

While this design choice enables deeper access to model internals, it also necessitates operating in a white-box setting, as the confidence model relies on activation signals from the LLM to assess correctness.  Furthermore, the method is customized to the specific architecture of the target model, meaning that adaptation to other LLMs may require reconfiguration and retraining.  
These limitations also present opportunities for future research: integrating confidence estimation directly into the generation process, reducing computational cost, and developing architecture-agnostic approaches that preserve the performance benefits of activation-based probing methods. 

A limitation of this study is that the dataset cannot be made publicly available. The data contains sensitive and proprietary information pertaining to internal financial tools and knowledge resources used by service agents within a financial institution. This restriction is mandated by internal data governance policies to protect confidential and regulated financial information.

\bibliography{custom}

\appendix
\end{document}